\documentclass[10pt,twocolumn,letterpaper]{article}

\usepackage{cvpr} 

\usepackage{graphicx}
\usepackage{amsmath}
\usepackage{amssymb}
\usepackage{booktabs}
\usepackage{array}
\usepackage{makecell}
\usepackage{cite}

\usepackage[pagebackref,breaklinks,colorlinks]{hyperref}

\usepackage[capitalize]{cleveref}
\crefname{section}{Sec.}{Secs.}
\Crefname{section}{Section}{Sections}
\Crefname{table}{Table}{Tables}
\crefname{table}{Tab.}{Tabs.}

\def\cvprPaperID{8798} 
\def\confName{CVPR}
\def\confYear{2023}

\begin{document}
\title{Learning Conditional Attributes for Compositional Zero-Shot Learning}
\author{Qingsheng Wang$^{1}$\footnotemark[1] ,
Lingqiao Liu$^2$,
Chenchen Jing$^3$,
Hao Chen$^3$,
Guoqiang Liang$^{1}$,\\
Peng Wang$^{1}$\footnotemark[2] ,
Chunhua Shen$^3$\\
$^1$School of Computer Science and Ningbo Institute, Northwestern Polytechnical University, Xi'an, China\\
$^2$School of Computer Science, University of Adelaide, Adelaide, Australia\\
$^3$School of Computer Science, Zhejiang University, Hangzhou, China\\
}
\maketitle
\renewcommand{\thefootnote}{\fnsymbol{footnote}}
\footnotetext[1]{E-mail: wqshmzh@mail.nwpu.edu.cn}
\footnotetext[2]{Corresponding author. E-mail: peng.wang@nwpu.edu.cn}
\footnotetext[3]{Gitee: \url{https://gitee.com/wqshmzh/canet-czsl}}

\setlength\abovedisplayskip{2pt}
\setlength\belowdisplayskip{2pt}
\setlength{\belowcaptionskip}{-0.35cm}

\begin{abstract}
Compositional Zero-Shot Learning (CZSL) aims to train models to recognize novel compositional concepts based on learned concepts such as attribute-object combinations. One of the challenges is to model attributes interacted with different objects, e.g., the attribute ``wet" in ``wet apple" and ``wet cat" is different. As a solution, we provide analysis and argue that attributes are conditioned on the recognized object and input image and explore learning conditional attribute embeddings by a proposed attribute learning framework containing an attribute hyper learner and an attribute base learner. By encoding conditional attributes, our model enables to generate flexible attribute embeddings for generalization from seen to unseen compositions. Experiments on CZSL benchmarks, including the more challenging C-GQA dataset, demonstrate better performances compared with other state-of-the-art approaches and validate the importance of learning conditional attributes. Code\footnotemark[3] is available at \url{https://github.com/wqshmzh/CANet-CZSL}.
\end{abstract}

\section{Introduction}
\label{Introduction}
Deep machine learning algorithms today can learn knowledge of concepts to recognize patterns. Can a machine compose different learned concepts to generalize to new compositions? Compositional generalization is one of the hallmarks of human intelligence \cite{atzmon2016learning, lake2017building}. To make the models equipped with this ability, Compositional Zero-Shot Learning (CZSL) \cite{misra2017red} is proposed, where the models are trained to recognize images of unseen compositions composed of seen concepts. In this work, we concentrate on the situation where each composition is composed by attribute (e.g., \emph{wet}) and object (e.g., \emph{apple}). For example, given images of \emph{wet apple} and \emph{dry cat}, a well-trained model can recognize images of new compositions \emph{dry apple} and \emph{wet cat}.

\begin{figure}[t]
\centering
\includegraphics[width=1\linewidth]{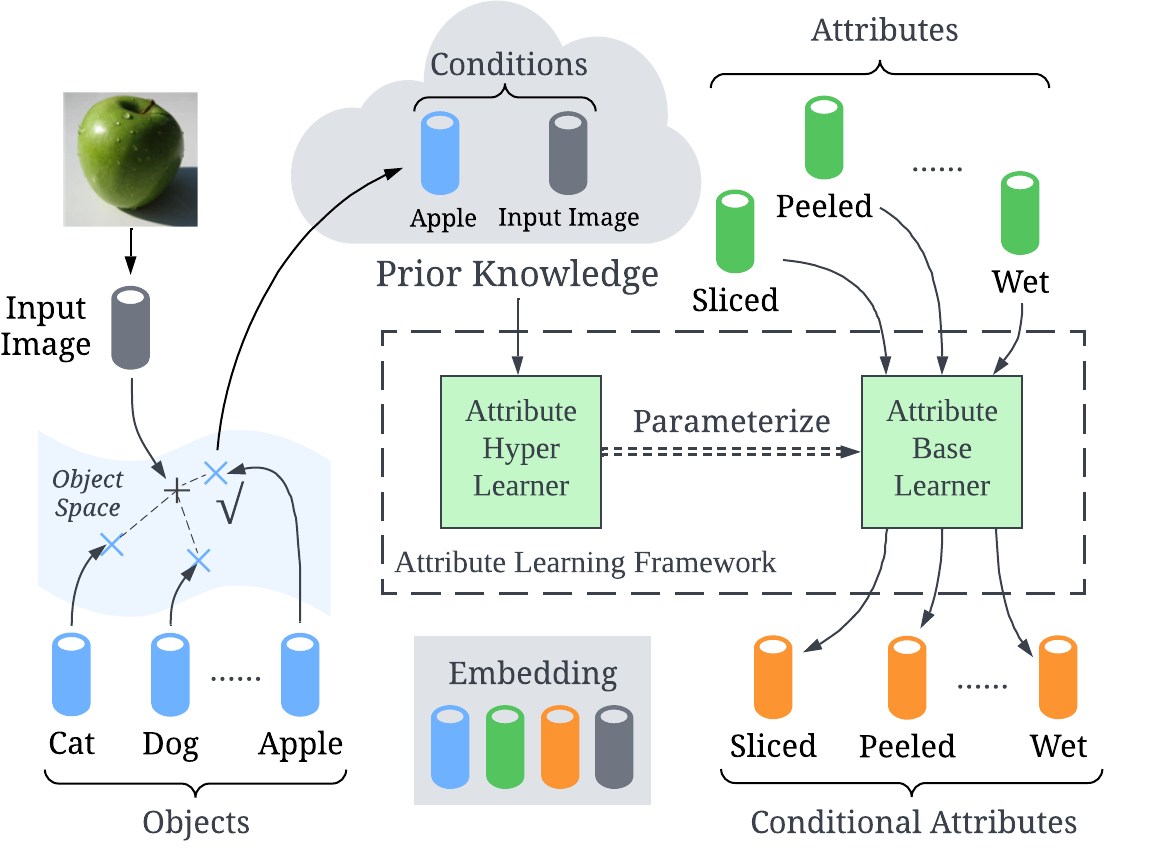}
\caption{The diagram of our work. We aim to learn conditional attributes conditioned on the recognized object and input image through an attribute learning framework containing an attribute hyper learner and an attribute base learner. We first recognize the object in the input image. Then, we feed prior knowledge extracted from the conditions, which are recognized object word embedding and input image visual embedding, to the attribute hyper learner. Finally, conditional attribute embeddings are produced by the attribute base learner parameterized by the attribute hyper learner.}
\label{fig1}
\end{figure}

Compositional Zero-Shot Learning of attribute-object compositions requires modeling attributes, objects, and the contextuality between them. Learning to model objects in CZSL is similar to conventional supervised object classification task since the model has access to all objects in CZSL task \cite{saini2022disentangling}. Learning to model contextuality between attribute and object is mostly addressed in the literature \cite{misra2017red, purushwalkam2019task, yang2020learning, xu2021relation, mancini2021open, xu2021zero, naeem2021learning}. One of the main challenges of CZSL is the appearance diversity of an attribute when composed with different objects, e.g., attribute \emph{wet} in \emph{wet apple} and \emph{wet cat} is different. This reveals that the information of each attribute is dependent on different objects. However, most recent works in CZSL \cite{2018attrasop, atzmon2020causal, ruis2021independent, saini2022disentangling, yang2022decomposable, zhang2022learning} extract attribute representations irrelevant to the object from seen compositions to infer the unseen compositions. These approaches neglect the nature of attribute diversity and learn concrete attribute representation agnostic to different objects.

In this paper, we learn conditional attributes rather than learning concrete ones in a proposed \textbf{C}onditional \textbf{A}ttribute \textbf{Net}work (CANet). We first conduct analysis to determine the exact conditions by considering the recognition of attribute and object as computing a classification probability of attribute and object conditioned on the input image. By decomposing this probability, we demonstrate that the probability of the input image belonging to an attribute is conditioned on the recognized object and the input image.

We present an attribute learning framework to learn conditional attribute embeddings conditioned on the above two conditions. The framework contains an attribute hyper learner and an attribute base learner, which are sketched in \cref{fig1}. The attribute hyper learner learns from prior knowledge extracted from the conditions. The attribute base learner is parameterized by the attribute hyper learner and is designed to encode all attribute word embeddings into conditional attribute embeddings. With the attribute learning framework, the attribute embeddings are changed along with the recognized object and input image. Finally, the attribute matching is processed in an attribute space where the input image embedding is projected. The attribute classification logits are computed by cosine similarities between the projected input image embedding and all conditional attribute embeddings. Additionally, we model the contextuality between attribute and object as composing attribute and object word embeddings. We use cosine similarities between the projected input image embedding and all composed attribute-object embeddings to get the classification logits.

Our main contributions are as follows:
\begin{itemize}
\vspace{-0.3cm}
\item We propose to learn attributes conditioned on the recognized object and input image.
\vspace{-0.3cm}
\item We propose an attribute learning framework containing an attribute hyper learner and an attribute base learner for learning conditional attribute embeddings.
\vspace{-0.3cm}
\item Experiments and ablation studies indicate the effectiveness of our proposed conditional attribute network, which further validates the importance of learning conditional attributes in the CZSL task.
\end{itemize}


\section{Related Work}
\textbf{Compositional Zero-Shot Learning.} Given descriptions only, we can recognize objects that are never seen before. In conventional Zero-Shot Learning (ZSL), models have access both to images of seen classes and descriptions of seen and unseen classes \cite{lampert2009learning}. In contrast, CZSL presents no description of seen and unseen attribute-object compositions while all attributes and objects as concepts are seen during training. Recently, works in CZSL are divided into two main streams. One extracts attribute and object words or visual features independently from a composition during training, including learning attributes as linear transformations of objects \cite{2018attrasop}, learning to hierarchically decompose compositions and recompose the concepts with learned visual concepts \cite{yang2020learning}, learning independent prototypes of attributes and objects and compositing prototypes via graph network \cite{ruis2021independent}, and learning decomposed prototypes of visual concept features \cite{saini2022disentangling} via siamese contrastive embedding network \cite{li2022siamese}. The other is to learn a compositional space \cite{mancini2021open}, a graph network \cite{naeem2021learning, anwaar2022leveraging}, an episode-based cross-attention module \cite{xu2021zero}, and a contrastive space \cite{anwaar2021contrastive} for contextuality modeling. Also, Yang \emph{et al.} \cite{yang2022decomposable} rethink the CZSL task in a decomposable causal way and learn three spaces for attribute, object, and composition classifications. Additionally, with pre-trained large vision language models like CLIP, Nayak \emph{et al.} \cite{nayak2022learning} propose to tune soft prompts as concept embeddings.

Recent work in \cite{huang2021translational} addresses the problem of attribute diversity. They propose to learn translational attribute features conditionally dependent on the object prototypes. Specifically, they add generic object embedding as the object prototype to the concatenated attribute and object embedding. However, this approach makes the model concentrate more on the composition instead of the attribute, causing the attribute learning degrade to learning the contextuality between attribute and object. On the contrary, we explicitly focus on learning conditional attribute embeddings. The learned conditional attribute embeddings can be changed along with the objects and input images.

\textbf{Attribute Learning.} Learning features of attributes is explored by a large community including image search \cite{kovashka2012whittlesearch, siddiquie2011image}, sentence generation \cite{kulkarni2013babytalk}, and zero-shot classification \cite{chen2020learning, nan2019recognizing}. Conventional attribute learning approaches map the attributes into high-dimensional space and train a discriminative classification head without considering the diverse nature of attributes \cite{singh2016end, su2016deep, lu2017fully}. Our work also learns high-dimensional embeddings to represent attributes. The main difference is that our learned attribute embeddings are conditioned on different objects and input images.


\begin{figure*}[t]
\centering
\includegraphics[width=1\linewidth]{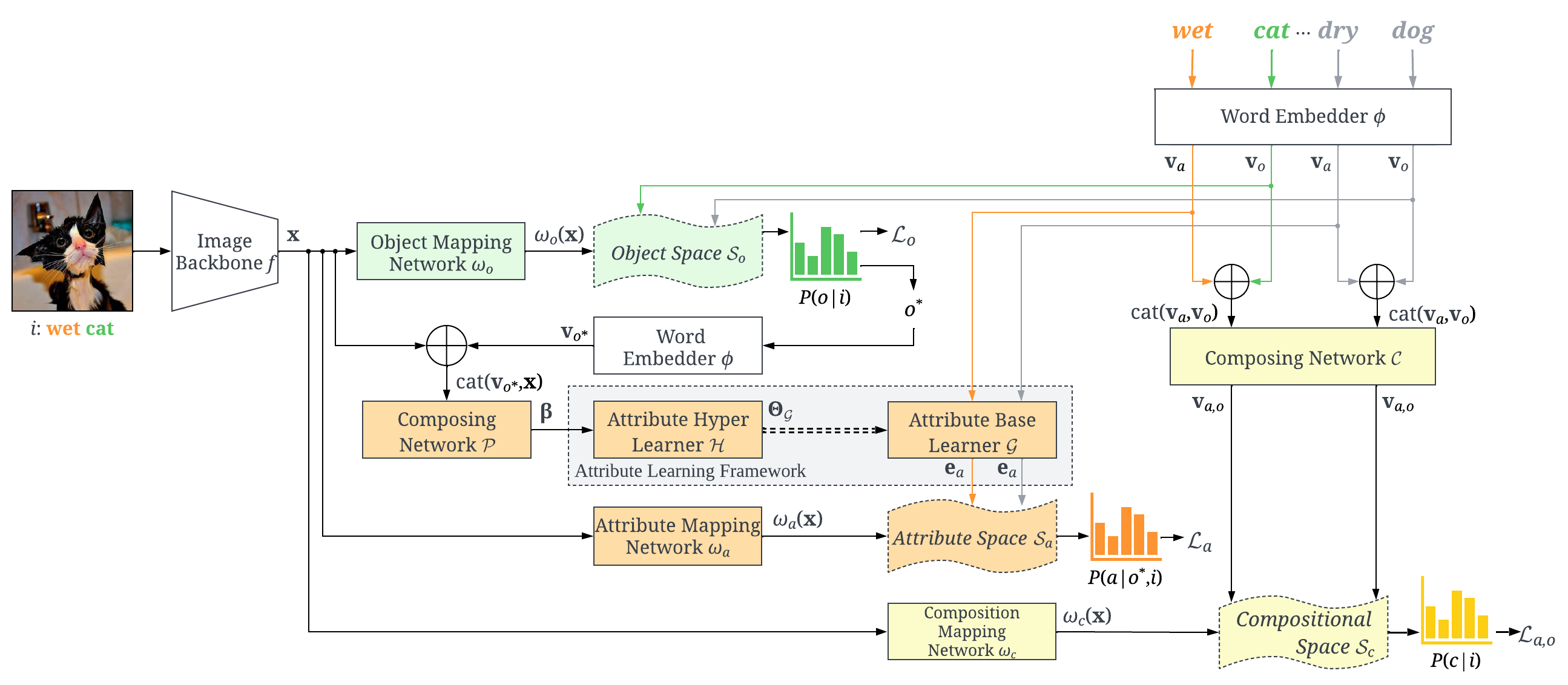}
\caption{Structure of our proposed CANet. The symbol $\bigoplus$ is channel-wise concatenation. The mapping networks $\mathcal{\omega}_o$, $\mathcal{\omega}_a$, and $\mathcal{\omega}_c$ map the input image embedding $\boldsymbol{\mathrm{x}}$ into object, attribute, and composition spaces $\mathcal{S}_o$, $\mathcal{S}_a$, and $\mathcal{S}_c$. All object word embeddings $\boldsymbol{\mathrm{v}}_o$ along with the object-mapped input image embedding are used in $\mathcal{S}_o$ to compute loss $\mathcal{L}_o$ and get the recognized object $o^*$. The attribute hyper learner $\mathcal{H}$ learns to parameterize the attribute base learner $\mathcal{G}$ using the prior knowledge $\boldsymbol{\mathrm{\beta}}$ extracted from the recognized object word embedding $\boldsymbol{\mathrm{v}}_{o^*}$ and $\boldsymbol{\mathrm{x}}$. The conditional attribute embeddings $\boldsymbol{\mathrm{e}}_a$ are encoded by $\mathcal{G}$ parameterized by $\mathcal{H}$. Using all $\boldsymbol{\mathrm{e}}_a$ along with the attribute-mapped input image embedding in space $\mathcal{S}_a$ to compute loss $\mathcal{L}_a$. Loss $\mathcal{L}_{a,o}$ is computed using all compositional word embeddings produced by composing network $\mathcal{C}$ and the composition-mapped input image embedding in space $\mathcal{S}_c$.}
\label{fig2}
\end{figure*}

\section{Approach}

\subsection{Task Definition}
The task of CZSL aims to learn to classify an image $i$ into a composition $c$ composed by multiple seen concepts, where $i$ and $c$ are unseen during training. Denote sets of images, compositions, attributes, and objects as $\mathcal{I}$, $\mathcal{C}$, $\mathcal{A}$, and $\mathcal{O}$, we have $i\in \mathcal{I}$, $c\in \mathcal{C}$, $a\in \mathcal{A}$, $o\in \mathcal{O}$, and $\mathcal{C}=\mathcal{A} \times \mathcal{O}$. During training, machines have access to seen set $\mathcal{D}_{seen}=\{(i^s,c^s)|i^s\in \mathcal{I}^{s}, \mathcal{I}^s\subsetneqq \mathcal{I},c\in \mathcal{C}^s,\mathcal{C}^s\subsetneqq \mathcal{C}\}$, attribute set $\mathcal{A}$, and object set $\mathcal{O}$, where $\mathcal{I}^s$ and $\mathcal{C}^s$ are sets of images and compositions seen when training. Also, evaluation of algorithms requires unseen set $\mathcal{D}_{unseen}=\{(i^u,c^u)|i^u\in \mathcal{I}^u,c^u\in \mathcal{C}^u,\mathcal{I}^u\subsetneqq \mathcal{I},\mathcal{C}^u\subsetneqq \mathcal{C}\}$ used for validation and testing. In conventional ZSL, $\mathcal{I}^s\cap \mathcal{I}^u=\varnothing$, $\mathcal{C}^s\cap \mathcal{C}^u=\varnothing$, i.e. unseen images and compositions are not overlapped with the seen ones. Here, we follow the setting of Generalized Zero-Shot Learning (GZSL) where images in $\mathcal{I}^s$ and $\mathcal{I}^u$ and labels in $\mathcal{C}^s$ and $\mathcal{C}^u$ appear during validation and testing. GZSL is a challenging setting with larger label space and a strong bias of seen compositions to unseen ones, relaxing the less realistic assumption in conventional ZSL that test data only belongs to unseen classes.

\subsection{Conditional Attribute Network}
\noindent\textbf{Determining Conditions.} In CZSL, it is common to learn to classify attributes and objects besides compositions. We first assume that the model recognizes the input image $i$ as attribute $a^*$ and object $o^*$. The recognizing score can be formulated as a conditional probability $P(a^*,o^*|i)$ conditioned on input image $i$. We propose to decompose this probability to express the attribute and object recognition as a one-label classification task. According to multi-variable conditional probability formulation, we have:
\begin{equation}
P(a^*,o^*|i)=P(a^*|o^*,i)P(o^*|i)
\label{eq1}
\end{equation}
where $P(o^*|i)$ is probability of image $i$ belonging to object $o^*$ and $P(a^*|o^*,i)$ represents the probability of attribute $a^*$ conditioned on the joint presentation of $o^*$ and $i$. This indicates that the recognition of an attribute is conditioned on the recognized object and input image. To better solve the attribute diversity problem, we consider the information of the recognized object and input image as conditions for conditional attribute encoding.

\noindent\textbf{Object Recognition.} Object recognition requires learning to map the input image embedding into an object space. To incorporate object semantic information, we compute cosine similarities between object-mapped image embedding $\omega_o(\boldsymbol{\mathrm{x}})$ and all object word embeddings $\boldsymbol{\mathrm{v}}_o$ instead of directly learning a classification head:
\begin{equation}
\langle\mathcal{\omega}_o(\boldsymbol{\mathrm{x}}),\boldsymbol{\mathrm{v}}_o\rangle=\frac{\mathcal{\omega}_o(\boldsymbol{\mathrm{x}})^\top\boldsymbol{\mathrm{v}}_o}{\Vert\mathcal{\omega}_o(\boldsymbol{\mathrm{x}})\Vert_2\Vert\boldsymbol{\mathrm{v}}_o\Vert_2}
\label{eq2}
\end{equation}
where $\boldsymbol{\mathrm{x}}=f(i)\in\mathcal{X}$ is the visual embedding of input image $i$ in visual space $\mathcal{X}$ extracted by image backbone $f$ (e.g., ResNet-18 \cite{he2016deep}), $\mathcal{\omega}_o$ is the mapping from $\mathcal{X}$ to object semantic space $\mathcal{S}_o$, $\boldsymbol{\mathrm{v}}_o=\phi(o)$, $\boldsymbol{\mathrm{v}}_o\in\mathcal{S}_o$ is object word embedding in $\mathcal{S}_o$ extracted by word embedder $\phi$ (e.g., word2vec \cite{mikolov2013distributed}, FastText \cite{bojanowski2017enriching}). The recognized object $o^*=\underset{o\in\mathcal{O}}{\arg\max}\langle\mathcal{\omega}_o(\boldsymbol{\mathrm{x}}),\phi(o)\rangle$.

\noindent\textbf{Learning Conditional Attributes.} With the recognized object $o^*$ and visual embedding $\boldsymbol{\mathrm{x}}$ of input image $i$, we learn the attribute hyper learner $\mathcal{H}$ and attribute base learner $\mathcal{G}$ in the proposed attribute learning framework to extract attribute embeddings conditioned on the recognized object $o^*$ and input image $i$. We consider the information of $o^*$ and $i$ as prior knowledge for $\mathcal{H}$. Specifically, the prior knowledge is implemented as a feature vector $\boldsymbol{\mathrm{\beta}}$:
\begin{equation}
\boldsymbol{\beta}=\mathcal{P}(\mathrm{cat}(\boldsymbol{\mathrm{v}}_{o^*},\boldsymbol{\mathrm{x}}))
\label{eq5}
\end{equation}
where $\boldsymbol{\mathrm{v}}_{o^*}=\phi(o^*)$ is word embedding of $o^*$, $\mathcal{P}$ is a composing network for $\boldsymbol{\mathrm{v}}_{o^*}$ and $\boldsymbol{\mathrm{x}}$, $\mathrm{cat(\cdot)}$ is channel-wise concatenation operation.
With the prior knowledge, the attribute hyper learner $\mathcal{H}$ can parameterize $\mathcal{G}$ as:
\begin{equation}
\boldsymbol{\Theta}_{\mathcal{G}}=\mathcal{H}(\boldsymbol{\mathrm{\beta}};\boldsymbol{\Theta}_{\mathcal{H}})
\end{equation}
where $\boldsymbol{\Theta}_{\mathcal{H}}$ is the set of randomly initialized parameters of $\mathcal{H}$, $\boldsymbol{\Theta}_{\mathcal{G}}$ is the set of generated parameters of $\mathcal{G}$. In this way, the attribute embeddings $\boldsymbol{\mathrm{e}}_a$ conditioned on $o^*$ and $i$ can be encoded via $\mathcal{G}$ parameterized by $\mathcal{H}$:
\begin{equation}
\boldsymbol{\mathrm{e}}_{a}=\mathcal{G}(\boldsymbol{\mathrm{v}}_{a};\boldsymbol{\Theta}_{\mathcal{G}})
\label{eq4}
\end{equation}
where $\boldsymbol{\mathrm{v}}_{a}=\phi(a)$ is word embedding of attribute word $a$.

\noindent\textbf{Modeling Contextuality.} Although the conditional attribute embeddings are related to the recognized object and input image, object embeddings are also supposed to be influenced by attributes. Therefore, we model the contextuality of attribute-object compositions to address the relationships between them. We follow the work of Mancini \emph{et al.} \cite{mancini2021open} in their closed world setting that contextuality is modeled as the mixture of attribute and object word embeddings to extract attribute-object compositional embeddings:
\begin{equation}
\boldsymbol{\mathrm{v}}_{a,o}=\mathcal{C}(\mathrm{cat}(\boldsymbol{\mathrm{v}}_a,\boldsymbol{\mathrm{v}}_o))
\end{equation}
where $\mathcal{C}$ is a composing network for $\boldsymbol{\mathrm{v}}_a$ and $\boldsymbol{\mathrm{v}}_o$.

The entire structure of our model is shown in \cref{fig2}.

\subsection{Training Objectives}
Similar to object recognition, attribute or composition recognition is also implemented by computing cosine similarities $\langle\omega_a({\boldsymbol{\mathrm{x}}}),\boldsymbol{\mathrm{v}}_a\rangle$ or $\langle\omega_c({\boldsymbol{\mathrm{x}}}),\boldsymbol{\mathrm{v}}_{a,o}\rangle$ between attribute-mapped or composition-mapped image embeddings, \emph{i.e.}, $\omega_a({\boldsymbol{\mathrm{x}}})$ or $\omega_c({\boldsymbol{\mathrm{x}}})$, and attribute word embeddings or attribute-object compositional embeddings, \emph{i.e.}, $\boldsymbol{\mathrm{v}}_a$ or $\boldsymbol{\mathrm{v}}_{a,o}$:
\begin{equation}
\langle\omega_a({\boldsymbol{\mathrm{x}}}),\boldsymbol{\mathrm{e}}_a\rangle=\frac{\mathcal{\omega}_a(\boldsymbol{\mathrm{x}})^\top\boldsymbol{\mathrm{e}}_a}{\Vert\mathcal{\omega}_a(\boldsymbol{\mathrm{x}})\Vert_2\Vert\boldsymbol{\mathrm{e}}_a\Vert_2}
\end{equation}
\begin{equation}
\langle\omega_c({\boldsymbol{\mathrm{x}}}),\boldsymbol{\mathrm{v}}_{a,o}\rangle=\frac{\mathcal{\omega}_c(\boldsymbol{\mathrm{x}})^\top\boldsymbol{\mathrm{v}}_{a,o}}{\Vert\mathcal{\omega}_c(\boldsymbol{\mathrm{x}})\Vert_2\Vert\boldsymbol{\mathrm{v}}_{a,o}\Vert_2}
\end{equation}
The recognition probability $P(o^*|i)$, $P(a^*|o^*,i)$, and $P(c^*|i)$ are normalized cosine similarities, where $P(c^*|i)$ is the probability of input image $i$ belonging to the recognized attribute-object composition $c^*$. As shown in \cref{fig2}, our model learns three embedding spaces: attribute space $\mathcal{S}_a$, object space $\mathcal{S}_o$, and attribute-object compositional space $\mathcal{S}_c$. Therefore, we incorporate three separate cross-entropy losses to maximize the three recognition probabilities to make the model optimized in these three spaces. The losses are as follows:
\begin{equation}
\mathcal{L}_a=-\sum_{a\in\mathcal{A}}\log\frac{\exp(\langle\mathcal{\omega}_a(\boldsymbol{\mathrm{x}}),\boldsymbol{\mathrm{e}}_a\rangle/\tau)}{\sum_{a'\in\mathcal{A}}\exp(\langle\mathcal{\omega}_a(\boldsymbol{\mathrm{x}}),\boldsymbol{\mathrm{e}}_{a'}\rangle/\tau)}
\end{equation}
\begin{equation}
\,\,\,\,\,\,\,\,\,\,\,\mathcal{L}_o=-\sum_{o\in\mathcal{O}}\log\frac{\exp(\langle\mathcal{\omega}_o(\boldsymbol{\mathrm{x}}),\boldsymbol{\mathrm{v}}_o\rangle/\tau)}{\sum_{o'\in\mathcal{O}}\exp(\langle\mathcal{\omega}_o(\boldsymbol{\mathrm{x}}),\boldsymbol{\mathrm{v}}_o')\rangle/\tau)}
\end{equation}
Also, for composition recognition, we have:
\begin{equation}
\mathcal{L}_{a,o}=-\sum_{(a,o)\in\mathcal{C}}\log\frac{\exp(\langle\mathcal{\omega}_c(\boldsymbol{\mathrm{x}}),\boldsymbol{\mathrm{v}}_{a,o}\rangle/\tau)}{\sum_{(a',o')\in\mathcal{C}}\exp(\langle\mathcal{\omega}_c(\boldsymbol{\mathrm{x}}),\boldsymbol{\mathrm{v}}_{a',o'}\rangle/\tau)}
\end{equation}
where $\tau$ is temperature factor \cite{zhang2019adacos}. Finally, the training loss as a whole linearly combines the three losses above:
\begin{equation}
\mathcal{L}=\frac{\mathcal{L}_a+\mathcal{L}_o}{2}+\mathcal{L}_{a,o}
\end{equation}

\subsection{Inference}
During validation and testing, we incorporate a linear normalization function $g$ for cosine similarities:
\begin{equation}
g(d)=(1+d)*0.5
\end{equation}
Then, we have $P(a|o^*,i)=g(\langle\mathcal{\omega}_a(\boldsymbol{\mathrm{x}}),\boldsymbol{\mathrm{e}}_a\rangle)$, $P(o|i)=g(\langle\mathcal{\omega}_o(\boldsymbol{\mathrm{x}}),\boldsymbol{\mathrm{v}}_o\rangle)$, and $P(c|i)=g(\langle\mathcal{\omega}_c(\boldsymbol{\mathrm{x}}),\boldsymbol{\mathrm{v}}_{a,o}\rangle)$. The inference rule is parameterized as:
\begin{equation}
s=(1-\alpha)P(c|i)+\alpha P(a|o^*,i)P(o|i)
\label{inference}
\end{equation}
where $\alpha$ is the weight factor controlling balance.

\begin{table*}[t]
\centering
\renewcommand{\arraystretch}{1.2}
\begin{tabular}{l p{0.6cm} cc c cc c ccc c ccc}
\Xhline{1pt}
& & & & & \multicolumn{2}{c}{\textbf{Training}} & & \multicolumn{3}{c}{\textbf{Validation}} & & \multicolumn{3}{c}{\textbf{Test}} \\
\textbf{Dataset} & & $\mathcal{A}$ & $\mathcal{O}$ & & $\mathcal{C}_s$ & $\mathcal{I}$ & & $\mathcal{C}_s$ & $\mathcal{C}_u$ & $\mathcal{I}$ & & $\mathcal{C}_s$ & $\mathcal{C}_u$ & $\mathcal{I}$\\
\Xhline{0.5pt}
UT-Zappos50K \cite{yu2014fine, yu2017semantic}
&&16	&12 &&83 &22998	&&15 &15	&3214	&&18	&18	&2914	\\
MIT-States \cite{isola2015discovering}
&&115	&245	&&1262	&30338	&&300 &300	&10420	&&400	&400	&12995	\\
C-GQA \cite{naeem2021learning}
&&413	&674	&&5592	&26920	&&1040	&1252	&7280	&&888	&923	&5098	\\
\Xhline{1pt}
\end{tabular}
\caption{Detailed dataset splits of UT-Zappos50K, MIT-States, and C-GQA in training, validation, and test sets.}
\label{tab1}
\end{table*}

\section{Experiments}
In this section, experiments are conducted following the concrete introductions of datasets, metrics, implementation details, and baselines. Then, we report ablation results to demonstrate the effectiveness of our model.
\subsection{Experimental Setup}
\textbf{Datasets} We conduct experiments with three widely adopted datasets in the CZSL task, which are MIT-States \cite{isola2015discovering}, UT-Zappos50K \cite{yu2014fine, yu2017semantic}, and C-GQA \cite{naeem2021learning}. MIT-States contains 53753 crawled web images labeled with 1962 attribute-object (e.g., \emph{mossy highway}) compositions. This dataset has 30338, 10420, and 12995 training, validation, and testing images \cite{purushwalkam2019task} labeled with 1262, 600, and 800 compositions. In validation and test sets, the numbers of seen and unseen compositions are the same. All compositions are composed of 115 attributes and 245 objects. UT-Zappos50K is made up of 50025 images labeled with 116 fine-grained shoe classes composed of 16 attributes (e.g., \emph{rubber}) and 12 objects (e.g., \emph{sneaker}). This dataset has 22998, 3214, and 2914 training, validation, and testing images \cite{purushwalkam2019task} labeled with 83, 30, and 36 compositions. Also, numbers of seen and unseen compositions in validation and test sets share the same quantity. C-GQA is created based on Stanford GQA dataset \cite{hudson2019gqa} used for VQA task. C-GQA contains 39298 images labeled with 7767 compositions composed of 413 attributes and 674 objects. This dataset has 26920, 7280, and 5098 training, validation, and testing images labeled with 5592, 2292, and 1811 compositions. Detailed splits are presented in \cref{tab1}.

\textbf{Metrics} To demonstrate the advances in attribute learning, we report the attribute and object classification accuracies (best attr and best obj). The setting of GZSL requires both seen and unseen compositions to exist during validation and testing. As a result, there is an inherent bias of seen against unseen compositions. We follow the evaluation protocols proposed in \cite{chao2016empirical} where a scalar bias is added to final activations of classes of seen compositions to calibrate the model. As the scalar varies from negative infinity to positive infinity, there must be a best operating point at which the bias between seen and unseen compositions is the lowest. We report the results in terms of the best accuracies of seen images (best seen), unseen images (best unseen), best Harmonic Mean (best HM), and Area Under Curve (AUC) with different scalar biases.

\textbf{Implementations} We consider image backbone $f$ as ResNet-18 pre-trained on ImageNet \cite{deng2009imagenet} to extract 512 dimension vectors following preceding works. The mapping networks $\mathcal{\omega}_a$, $\mathcal{\omega}_o$, and $\mathcal{\omega}_c$ share the similar structure of two Fully Connected (FC) layers with ReLU \cite{nair2010rectified}, LayerNorm \cite{ba2016layer}, and Dropout \cite{srivastava2014dropout} following the first FC layer. We adopt word embedder $\phi$ as 600-dimensional word2vec+FastText for MIT-States, 300-dimensional FastText for UT-Zappos50K, and word2vec for C-GQA. The layer structures of $\mathcal{G}$, $\mathcal{P}$, and $\mathcal{C}$ are the same as the mapping networks, where ReLU is added in $\mathcal{P}$ and $\mathcal{C}$ to the last FC layer. Weight generation for the attribute base learner $\mathcal{G}$ through the attribute hyper learner $\mathcal{H}$ requires more parameters and makes learning difficult, as noted by Bertinetto \emph{et al.} \cite{bertinetto2016learning}. Therefore, we adopt weight factorization in \cite{wang2019tafe} to reduce parameters for the attribute hyper learner $\mathcal{H}$, that is
\begin{equation}
\boldsymbol{\mathrm{e}}_a^{(i)}=(\boldsymbol{\mathrm{v}}_a\boldsymbol{\mathrm{W}}_{\mathcal{G}}^{(i)}+\boldsymbol{\mathrm{b}}_{\mathcal{G}}^{(i)})\odot\boldsymbol{\mathrm{\lambda}}_{\mathcal{G}}^{(i)}
\end{equation}
\vspace{-10pt}
\begin{equation}
\boldsymbol{\mathrm{\lambda}}_{\mathcal{G}}^{(i)}=\mathcal{H}(\boldsymbol{\mathrm{\beta}};\boldsymbol{\Theta}_{\mathcal{H}}^{(i)})
\vspace{-6.5pt}
\end{equation}
where $(i)$ indicates $i$th FC layer, $\boldsymbol{\mathrm{W}}_{\mathcal{G}}^{(i)}$ and $\boldsymbol{\mathrm{b}}_\mathcal{G}^{(i)}$ are the weight matrix and bias vector of $i$th FC layer in $\mathcal{G}$, $\{\boldsymbol{\mathrm{W}}_{\mathcal{G}}^{(i)},\boldsymbol{\mathrm{b}}_\mathcal{G}^{(i)}\}\subset\boldsymbol{\Theta}_\mathcal{G}$, $\boldsymbol{\Theta}_{\mathcal{H}}^{(i)}\subset\boldsymbol{\Theta}_{\mathcal{H}}$, and $\odot$ denotes element-wise multiplication. During training, we fix the image backbone $f$ and train other modules using Adam \cite{kingma2014adam} optimizer and an Nvidia GeForce GTX 1080Ti GPU with a learning rate and weight decay of 0.00005. The batch size is 256. The temperature factor $\tau$, weight factor $\alpha$, and the maximum number of epochs are set to 0.02, 0.4, and 500 for UT-Zappos50K; 0.05, 0.2, and 800 for MIT-States; and 0.05, 0.4, and 1000 for C-GQA.

\textbf{Baselines} We conduct experiments with the following algorithms:
1) \emph{AttrAsOp} \cite{2018attrasop} treats attributes as linear transformations on object vectors instead of data points in some high-dimensional space and optimizes the transformations through several regularizers in the loss function.
2) \emph{TMN} \cite{purushwalkam2019task} constructs task-driven modular networks in semantic space configured through a gating function conditioned on the task.
3) \emph{SymNet} \cite{li2020symmetry} proposes symmetry property in attribute-object compositions and group axioms as objectives in an end-to-end manner.
4) \emph{CGE$_{\mathrm{ff}}$} \cite{naeem2021learning} exploits dependencies between attributes, objects, and compositions through an end-to-end graph formulation where "ff" means fixed image feature backbone.
5) \emph{CompCos} \cite{mancini2021open} learns a mapping from image features to semantic space of compositions and computes cosine similarities between them.
6) \emph{DeCa} \cite{yang2022decomposable} rethinks the CZSL task in a decomposable causal perspective and learns three independent mappings from image feature space to attribute, object, and composition semantic space. Cosine similarities are also adopted.
7) \emph{SCEN} \cite{li2022siamese} computes visual prototypes of attributes and objects in a siamese contrastive space and proposes a designed State Transition Module to increase the diversity of training compositions.

\begin{table*}[t]
\centering
\renewcommand{\arraystretch}{1.2}
\setlength{\tabcolsep}{0.5mm}{}
\begin{tabular}{lc p{0.1cm} cccccc p{0.1cm} cccccc p{0.1cm} cccccc}
\Xhline{1pt}
\multicolumn{2}{c}{\textbf{Algorithm}} & & \multicolumn{6}{c}{\textbf{UT-Zappos50K}} & & \multicolumn{6}{c}{\textbf{MIT-States}} & & \multicolumn{6}{c}{\textbf{C-GQA}} \\
\makecell[c]{Name} & Venue & & Att. & Obj. & S. & U. & HM & AUC & & Att. & Obj. & S. & U. & HM & AUC & & Att. & Obj. & S. & U. & HM & AUC\\
\Xhline{0.5pt}
AttrAsOp \cite{2018attrasop}
&	ECCV'18	&
&	38.9 &	69.6 &	59.8 &	54.2 &	40.8 &	25.9	&
&	21.1 &	23.6 &	14.3 &	17.4 &	9.9 &	1.6 &
&	8.3 &	12.5 &	11.8 &	3.9 &	2.9 &	0.3 \\
TMN \cite{purushwalkam2019task}
&	ICCV'19	&
&	40.8 &	69.9 &	58.7 &	60.0 &	45.0 &	29.3	&
&	23.3 &	26.5 &	20.2 &	20.1 &	13.0 &	2.9 &
&	9.7 &	20.5 &	21.6 &	6.3 &	7.7 &	1.1 \\
SymNet \cite{li2020symmetry}
&	CVPR'20	&
&	41.3 &	68.6 &	49.8 &	57.4 &	40.4 &	23.4	&
&	26.3 &	28.3 &	24.4 &	25.2 &	16.1 &	3.0 &
&	15.0 &	23.1 &	27.0 &	10.8 &	10.9 &	2.2 \\
CGE$_{\mathrm{ff}}$ \cite{naeem2021learning}
&	CVPR'21	&
&	45.0 &	73.9 &	56.8 &	63.6 &	41.2 &	26.4	&
&	27.9 &	32.0 &	28.7 &	25.3 &	17.2 &	5.1 &
&	12.7 &	26.9 &	27.5 &	11.7 &	11.9 &	2.5 \\
CompCos \cite{mancini2021open}
&	CVPR'21	&
&	44.7 &	73.5 &	59.8 &	62.5 &	43.5 &	28.7	&
&	27.9 &	31.8 &	25.3 &	24.6 &	16.4 &	4.5 &
&	- &	- &	- &	- &	- &	- \\
DeCa \cite{yang2022decomposable}
&	TMM'22	&
&	- &	- &	62.7 &	63.1 &	46.3 &	31.6	&
&	- &	- &	29.8 &	25.2 &	18.2 &	5.3 &
&	- &	- &	- &	- &	- &	- \\
SCEN \cite{li2022siamese}
&	CVPR'22	&
&	47.3	&\textbf{75.6}	&\textbf{63.5}	&63.1	&\textbf{47.8}	&32.0	&
&	28.2	&32.2 &\textbf{29.9}	&25.2	&\textbf{18.4}	&5.3	&
&	13.6	&\textbf{27.9}	&28.9 &12.1	&12.4 &2.9	\\
\multicolumn{2}{c}{\textbf{Ours}} &
&\textbf{48.4}	&72.6 &61.0 &\textbf{66.3} &47.3 &\textbf{33.1}	&
&\textbf{30.2}	&\textbf{32.6}	&29.0 &\textbf{26.2} &17.9 &\textbf{5.4}	&
&\textbf{17.5}	&22.3 &\textbf{30.0}	&\textbf{13.2} &\textbf{14.5}	&\textbf{3.3}	\\
\Xhline{1pt}
\end{tabular}
\caption{Quantitive results on test sets of all datasets with the state-of-the-art in terms of best attr (Att.), best obj (Obj.), best seen (S.), best unseen (U.), best HM (HM), and AUC.}
\label{tab2}
\end{table*}

\subsection {Quantitative Analysis}
\label{sec4.3}
All results are computed on test sets of all datasets and from their published papers and \cite{naeem2021learning} for a fair comparison. We report quantitative results with the best AUC in \cref{tab2}.

From Tab.2, our model outperforms other state-of-the-art algorithms in terms of best attr, best unseen, and AUC in all three datasets including the recently proposed C-GQA, indicating the better attribute learning performance and generalization ability from seen to unseen compositions. Specifically, our model performs much better on C-GQA with more state-of-the-art results although it is a much more challenging dataset than MIT-States and UT-Zappos50K.

For UT-Zappos50K, the observations are that our model boosts attribute recognition accuracy, unseen image classification accuracy, and AUC from 47.3\%, 63.1\%, and 32.0\% of SCEN to the new state-of-the-art of 48.4\%, 66.3\%, and 33.1\% with 1.1\%, 3.2\%, and 1.1\% improvement respectively. For MIT-States, our model achieves 30.2\%, 32.6\%, 26.2\%, and 5.4\% accuracies for attribute and object classification, unseen image classification, and AUC on the test set, providing 2.0\%, 0.4\%, 1.0\%, and 0.1\% improvements on the recently proposed SCEN as the new state-of-the-art results. This indicates that the proposed conditional attribute network can truly improve the attribute recognition performance and consequently the unseen image classification and AUC.

For the more challenging dataset C-GQA, since it is significantly harder than MIT-States and UT-Zappos50K with 4.4$\times$ and 0.9$\times$ composition labels and images in the training set compared with MIT-States, our model outperforms all other algorithms except best obj with 3.9\%, 1.1\%, 1.1\%, 2.1\%, and 0.4\% boosting in terms of best attr, best seen, best unseen, best HM, and AUC in the testing set. This indicates that the proposed conditional attribute network makes a critical contribution when facing a more challenging dataset even if the object recognition accuracy is lower.

We give an analysis of the importance that attributes should be conditioned on objects. First, note that although DeCa also learns attribute, object, and composition spaces separately, it learns attributes as static embeddings independent from objects, causing lower best unseen and AUC on UT-Zappos50K and MIT-States compared with ours in \cref{tab2}. Next, different baselines incorporate different techniques to handle the CZSL task though, they learn static attribute embeddings too, producing lower best attr, best unseen, and AUC. Then, the proposed method performs much better on C-GQA compared with other baselines. All the above phenomena demonstrate that attributes should be conditioned on objects and performance on datasets with larger label space can gain more boosts in this way.

From the results of the three datasets above, we observe an interesting phenomenon. Although the results of best obj in three datasets are all lower than that of SCEN, results of best unseen are all higher accompanied by higher results of best attr. We speculate the reason is that some objects have more attributes (or are more dominated) in unseen compositions and the misclassified objects recognized by our model are less dominated (i.e. have few attributes or are long-tailed in terms of attribute). As a result, with the correctly predicted objects dominated in unseen compositions, the more correctly classified attributes, the higher the results of best unseen.

\begin{table*}[t]
\centering
\renewcommand{\arraystretch}{1.2}
\setlength{\tabcolsep}{1.6mm}{}
\begin{tabular}{cl p{1cm} cccccc p{1cm} cccccc}
\Xhline{1pt}
\multicolumn{2}{c}{\textbf{Variant}} & & \multicolumn{6}{c}{\textbf{UT-Zappos50K}} & & \multicolumn{6}{c}{\textbf{C-GQA}} \\
\# & \makecell[c]{Name} & & Att. & Obj. & S. & U. & HM & AUC & & Att. & Obj. & S. & U. & HM & AUC \\
\Xhline{0.5pt}
(1) & w/o $\mathcal{L}_a$ + $\mathcal{L}_o$	&
&	46.1 &	\textbf{74.3}	&	61.5 &	64.7 &	46.1 &	31.7 &
&	10.8 &	\textbf{30.6}	&	29.8 &	12.8 &	13.4 &	3.0 \\
(2) & w/o $\mathcal{L}_c$ &
&	41.9 &	60.7 &	59.5 &	54.7 &	45.7 &	28.3 &
&	14.8 &	17.1 &	28.1 &	11.2 &	12.4 &	2.6 \\
(3) & w/o $\mathcal{G}$ + $\mathcal{H}$ + $\mathcal{P}$ &
&	47.0 &	74.0 &	59.9 &	65.8 &	46.3 &	31.7 &
&	14.8 &	27.8 &	29.9 &	13.1 &	14.6 &	3.1 \\
(4) & w/o $\mathcal{P}$ &
&	46.7 &	73.2 &	60.7 &	64.5 &	47.5 &	31.6 &
&	14.5 &	26.4 &	30.1 &	13.0 &	14.5 &	3.1 \\
(5) & w/o $\mathcal{H}$ &
&	46.2 &	70.3 &	58.5 &	62.7 &	46.3 &	30.5 &
&	13.9 &	25.8 &	29.1 &	11.2 &	12.5 &	2.4 \\
(6) & w/o $\boldsymbol{\mathrm{x}}$ for $\mathcal{H}$ &
&	45.6 &	71.3 &	\textbf{61.6}&	62.8 &	44.8 &	30.1 &
&	13.4 &	20.8 &	30.2 &	12.7 &	13.9 &	2.9 \\
(7) & \textbf{Full} &
&	\textbf{48.4}	&	72.6 &	61.0 &	\textbf{66.3}	&	\textbf{47.3}	&	\textbf{33.1} &
&	\textbf{17.5}	&	22.3 &	\textbf{30.0}	&	\textbf{13.2}	&	\textbf{14.5}	&	\textbf{3.3}	\\
\Xhline{1pt}
\end{tabular}
\caption{Ablation results on test sets in terms of best attr (Att.), best obj (Obj.), best seen (S.), best unseen (U.), best HM (HM), and AUC.}
\label{tab3}
\end{table*}

\begin{figure*}[t]
\centering
\includegraphics[width=1\linewidth]{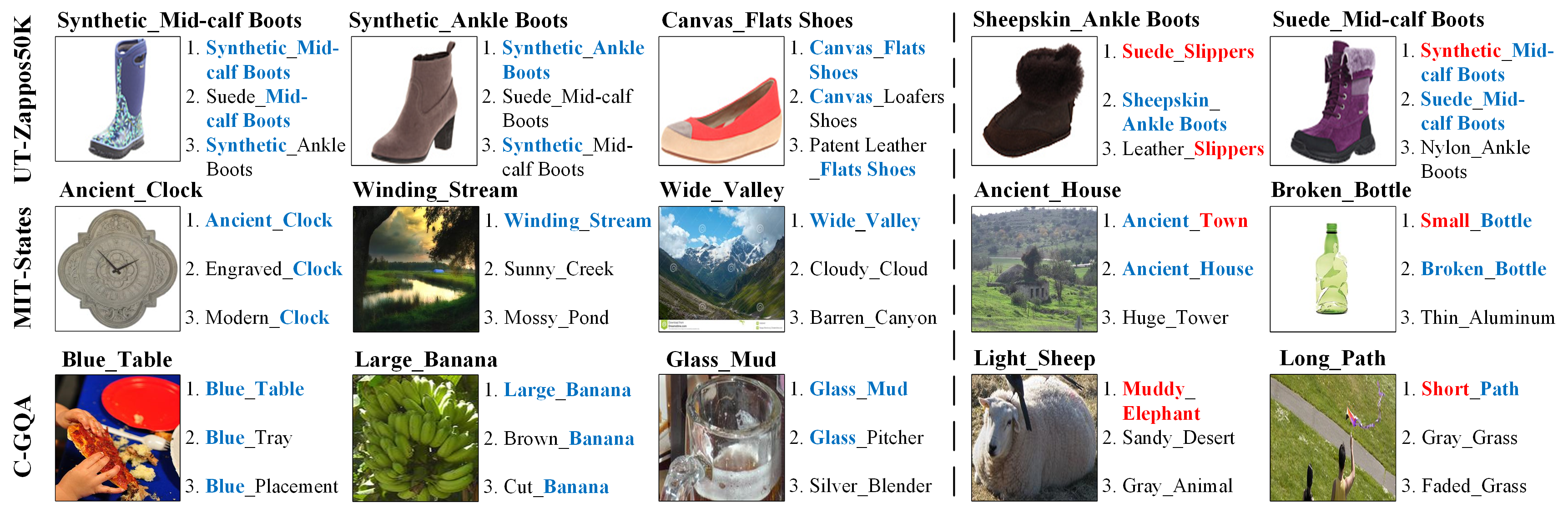}
\caption{Qualitative Results. We demonstrate top-3 predictions of some instances using our proposed model.}
\label{fig3}
\end{figure*}

\subsection{Qualitative Analysis}
In this section, we present some qualitative results of novel compositions with top-3 predictions on UT-Zappos50K, MIT-States, and C-GQA in \cref{fig3}. We show results for each dataset in each row. Images whose top prediction matches the label are shown in the first three columns and the rest columns show wrong results. For UT-Zappos50K, the remaining two answers of all images can match at least one label factor. For some instances in MIT-States, we can notice that the top and second predictions can both describe the image. For example, for the image labeled with \emph{winding stream}, there is sunlight reflecting from the stream and creek is the synonym of stream. Therefore, \emph{sunny creek} can also be the label of the image. Another example is that image labeled with \emph{wide valley} also present \emph{cloudy cloud} in the blue sky located in the upper part. As a result, the model has difficulty deciding what to predict. This reflects that labels in MIT-States are heavy in noise. For C-GQA where labels are clear, our model can produce more answers that match the label factors in the remaining two predictions, which indicates the better performance and robustness of our model.

Additionally, we present wrong predictions in the last two columns. It can be noticed that our model can predict correct answers in most top-3 predictions. As for the image of MIT-States labeled as \emph{broken bottle}, our model predicts the attribute as \emph{small} because it is difficult to focus on a certain attribute of the bottle since the bottle in this image is both small and broken. Besides, images in the training set are limited in the status of objects. For example, training images of \emph{sheep} hardly present the action of laying down causing our model mistakenly classify the sheep in the image labeled as \emph{light sheep} into \emph{elephant}. Thus, with the recognized object \emph{elephant}, the model can only focus on attributes conditioned on \emph{elephant} and find the appropriate attribute that matches the image.

\begin{figure*}[t]
\centering
\includegraphics[width=1\linewidth]{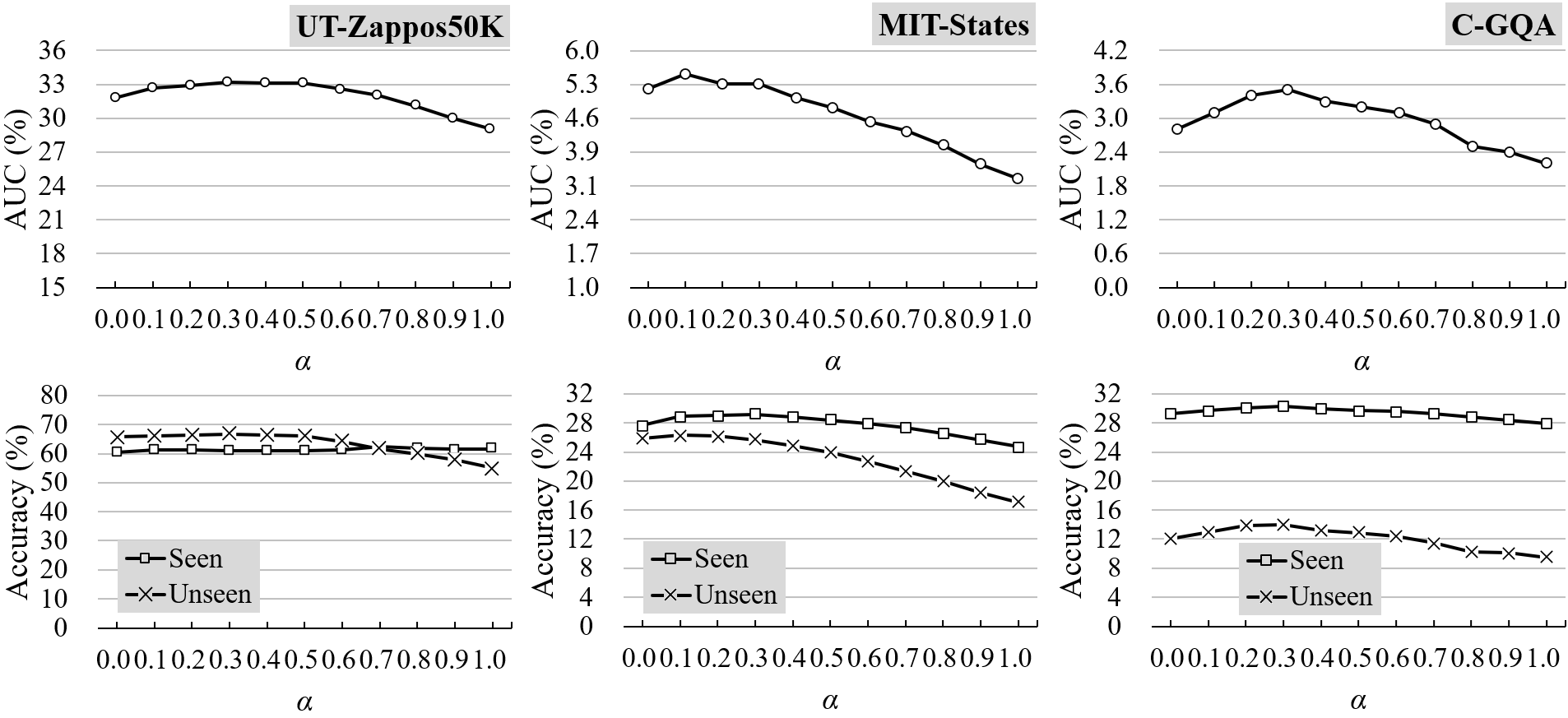}
\caption{Influence of weighting factor $\alpha$ on AUC, seen accuracy, and unseen accuracy.}
\label{fig4}
\end{figure*}

\subsection{Ablation Study}
In this section, we ablate the proposed model to evaluate the performance of each module. The ablation study is conducted on test sets of UT-Zappos50K, C-GQA. To achieve more convincing ablation results, we do not choose MIT-States because images in MIT-States are labeled using automatic search. As a result, the noise of labels is too heavy to be used for evaluations, as is pointed out by Atzmon et.al. \cite{atzmon2020causal}. Ablation results are reported in \cref{tab3}. The detailed ablation process is as follows:

We first study the effects of recognizing compositions only and recognizing attributes and objects without compositions, which are corresponded to variants (1) and (2) in \cref{tab3}, and report results in terms of six metrics used in \cref{tab2}. Compared with variant (6), the results of each variant mostly decline, indicating the importance of recognizing attributes, objects, and compositions jointly. As to the results of Obj. in UT-Zappos50K and C-GQA from variants (1) and (6), we can observe that those object classification accuracies decline. This is mainly because when recognizing objects individually each object presents visual diversity caused by different attributes, e.g., \emph{sliced apple} is different from \emph{apple} in visual appearance since a sliced apple is sliced into multiple pieces. However, this phenomenon does not affect other results, especially for the dataset with larger label space, e.g., C-GQA.

Next, with attributes, objects, and compositions being recognized, we ablate $\mathcal{G}$ + $\mathcal{H}$ + $\mathcal{P}$ (where attribute word embeddings are directly used in $\mathcal{S}_a$), $\mathcal{H}$ (where the concatenation of $\beta$ and $\boldsymbol{\mathrm{v}}_a$ is fed to $\mathcal{G}$), $\mathcal{P}$ (where the concatenation of $\boldsymbol{\mathrm{v}}_{o^*}$ and $\boldsymbol{\mathrm{x}}$ is directly fed to $\mathcal{H}$), and $\boldsymbol{\mathrm{x}}$ in $\boldsymbol{\mathrm{\beta}}$ that is fed to $\mathcal{H}$, which are corresponded to variants (3)-(6) in \cref{tab3}. Note that if $\mathcal{P}$ is presented then $\mathcal{G}$ is required to be presented also because the concatenation of $\beta$ and $\boldsymbol{\mathrm{v}}_a$ requires changing the number of dimensions to that of $\boldsymbol{\mathrm{x}}$ through $\mathcal{G}$. Also, $\mathcal{G}$ can not be ablated only otherwise the model changes to variant (3). Compared to variant (7), all results are mostly declined, indicating that optimizing the model with $\mathcal{G}$, $\mathcal{H}$, $\mathcal{P}$, and $\boldsymbol{\mathrm{x}}$ is essential. As for variant (5), it performs worse on seen and unseen images compared with variant (4), indicating the importance of using the attribute hyper learner. Comparing results of variants (5) with (6), we observe that adding the attribute hyper learner $\mathcal{H}$ for $\mathcal{G}$ without image visual embedding $\boldsymbol{\mathrm{x}}$ marginally increases overfitting to seen images. Results of variant (6) present the best generalization compared with variants (3)-(5) because adding $\boldsymbol{\mathrm{x}}$ in $\boldsymbol{\mathrm{\beta}}$ for $\mathcal{H}$ provides diverse instances that are seen for $\mathcal{H}$ during training since the number of image visual embeddings is far more than the number of object word embeddings.

Lastly, we conduct experiments to study the impact of weighting factor $\alpha$ in \cref{inference} on all datasets. In detail, we present results of AUC, seen accuracy, and unseen accuracy in \cref{fig4} with $\alpha\in [0,1]$ with an interval of 0.1. It can be observed that all results of AUC increase first and then decrease with $\alpha$ increases from 0.0. The peaks of AUC are reached when $\alpha$ equals 0.5, 0.1, and 0.3 in UT-Zappos50K, MIT-States, and C-GQA. This phenomenon reveals that learning to classify attributes, objects, and compositions all contribute to making our model reach optimal. Additionally, it can be noticed that the results of AUC are all generally declined when $\alpha$ changes from 0.0 to 1.0. This is because attribute-object compositions involve not only the side information of attribute and object, but also the contextuality between attribute and object. As for the results of seen and unseen accuracies reported in the second row, the same trend can be observed. Specifically, results of unseen accuracy are more sensitive than seen accuracy and also have peaks with various $\alpha$, indicating that $\alpha$ also has an impact on the generalization ability of our model. In conclusion, using a small weighting factor $\alpha$ is always better than using a larger one, indicating that modeling contextuality between attribute and object is a bit more beneficial to the CZSL task than separately classifying attributes and objects.

\section{Conclusion}
In this work, we address the attribute diversity problem in Compositional Zero-Shot Learning. As a solution, we propose a Conditional Attribute Network (CANet) to learn attributes conditioned on the recognized object and input image. We first decompose the probability of attribute and object recognition conditioned on the input image to lay a foundation for learning conditional attributes. Then, we build an attribute learning framework to encode conditional attribute embeddings. Experiments show that our model outperforms recent CZSL approaches and achieves new state-of-the-art results. Despite the better attribute recognition performance, a limitation is that our model is less qualified to handle object long-tailed distribution in terms of attribute mentioned in \cref{sec4.3}. Future works can be focused on solving the problem above while learning conditional attributes.

\section{Acknowledgements}
This work was supported by National Key R\&D Program of China (No.2020AAA0106900), the National Natural Science Foundation of China (No.U19B2037), Shaanxi Provincial Key R\&D Program (No.2021KWZ-03), and Natural Science Basic Research Program of Shaanxi (No.2021JCW-03).

\clearpage

{\small
\bibliographystyle{ieee_fullname}
\bibliography{ref}

\begin{thebibliography}{10}\itemsep=-1pt

\bibitem{anwaar2021contrastive}
Muhammad~Umer Anwaar, Rayyan~Ahmad Khan, Zhihui Pan, and Martin Kleinsteuber.
\newblock A contrastive learning approach for compositional zero-shot learning.
\newblock In {\em Proceedings of the 2021 International Conference on
  Multimodal Interaction}, pages 34--42, 2021.

\bibitem{anwaar2022leveraging}
Muhammad~Umer Anwaar, Zhihui Pan, and Martin Kleinsteuber.
\newblock On leveraging variational graph embeddings for open world
  compositional zero-shot learning.
\newblock {\em arXiv preprint arXiv:2204.11848}, 2022.

\bibitem{atzmon2016learning}
Yuval Atzmon, Jonathan Berant, Vahid Kezami, Amir Globerson, and Gal Chechik.
\newblock Learning to generalize to new compositions in image understanding.
\newblock {\em arXiv preprint arXiv:1608.07639}, 2016.

\bibitem{atzmon2020causal}
Yuval Atzmon, Felix Kreuk, Uri Shalit, and Gal Chechik.
\newblock A causal view of compositional zero-shot recognition.
\newblock {\em Advances in Neural Information Processing Systems},
  33:1462--1473, 2020.

\bibitem{ba2016layer}
Jimmy~Lei Ba, Jamie~Ryan Kiros, and Geoffrey~E Hinton.
\newblock Layer normalization.
\newblock {\em arXiv preprint arXiv:1607.06450}, 2016.

\bibitem{bertinetto2016learning}
Luca Bertinetto, Jo{\~a}o~F Henriques, Jack Valmadre, Philip Torr, and Andrea
  Vedaldi.
\newblock Learning feed-forward one-shot learners.
\newblock {\em Advances in neural information processing systems}, 29, 2016.

\bibitem{bojanowski2017enriching}
Piotr Bojanowski, Edouard Grave, Armand Joulin, and Tomas Mikolov.
\newblock Enriching word vectors with subword information.
\newblock {\em Transactions of the association for computational linguistics},
  5:135--146, 2017.

\bibitem{chao2016empirical}
Wei-Lun Chao, Soravit Changpinyo, Boqing Gong, and Fei Sha.
\newblock An empirical study and analysis of generalized zero-shot learning for
  object recognition in the wild.
\newblock In {\em European conference on computer vision}, pages 52--68.
  Springer, 2016.

\bibitem{chen2020learning}
Hui Chen, Zhixiong Nan, Jingjing Jiang, and Nanning Zheng.
\newblock Learning to infer unseen attribute-object compositions.
\newblock {\em arXiv preprint arXiv:2010.14343}, 2020.

\bibitem{deng2009imagenet}
Jia Deng, Wei Dong, Richard Socher, Li-Jia Li, Kai Li, and Li Fei-Fei.
\newblock Imagenet: A large-scale hierarchical image database.
\newblock In {\em 2009 IEEE conference on computer vision and pattern
  recognition}, pages 248--255. Ieee, 2009.

\bibitem{he2016deep}
Kaiming He, Xiangyu Zhang, Shaoqing Ren, and Jian Sun.
\newblock Deep residual learning for image recognition.
\newblock In {\em Proceedings of the IEEE conference on computer vision and
  pattern recognition}, pages 770--778, 2016.

\bibitem{huang2021translational}
He Huang, Wei Tang, Jiawei Zhang, and Philip~S Yu.
\newblock Translational concept embedding for generalized compositional
  zero-shot learning.
\newblock {\em arXiv preprint arXiv:2112.10871}, 2021.

\bibitem{hudson2019gqa}
Drew~A Hudson and Christopher~D Manning.
\newblock Gqa: A new dataset for real-world visual reasoning and compositional
  question answering.
\newblock In {\em Proceedings of the IEEE/CVF conference on computer vision and
  pattern recognition}, pages 6700--6709, 2019.

\bibitem{isola2015discovering}
Phillip Isola, Joseph~J Lim, and Edward~H Adelson.
\newblock Discovering states and transformations in image collections.
\newblock In {\em Proceedings of the IEEE conference on computer vision and
  pattern recognition}, pages 1383--1391, 2015.

\bibitem{kingma2014adam}
Diederik~P Kingma and Jimmy Ba.
\newblock Adam: A method for stochastic optimization.
\newblock {\em arXiv preprint arXiv:1412.6980}, 2014.

\bibitem{kovashka2012whittlesearch}
Adriana Kovashka, Devi Parikh, and Kristen Grauman.
\newblock Whittlesearch: Image search with relative attribute feedback.
\newblock In {\em 2012 IEEE Conference on Computer Vision and Pattern
  Recognition}, pages 2973--2980. IEEE, 2012.

\bibitem{kulkarni2013babytalk}
Girish Kulkarni, Visruth Premraj, Vicente Ordonez, Sagnik Dhar, Siming Li,
  Yejin Choi, Alexander~C Berg, and Tamara~L Berg.
\newblock Babytalk: Understanding and generating simple image descriptions.
\newblock {\em IEEE transactions on pattern analysis and machine intelligence},
  35(12):2891--2903, 2013.

\bibitem{lake2017building}
Brenden~M Lake, Tomer~D Ullman, Joshua~B Tenenbaum, and Samuel~J Gershman.
\newblock Building machines that learn and think like people.
\newblock {\em Behavioral and brain sciences}, 40, 2017.

\bibitem{lampert2009learning}
Christoph~H Lampert, Hannes Nickisch, and Stefan Harmeling.
\newblock Learning to detect unseen object classes by between-class attribute
  transfer.
\newblock In {\em 2009 IEEE conference on computer vision and pattern
  recognition}, pages 951--958. IEEE, 2009.

\bibitem{li2022siamese}
Xiangyu Li, Xu Yang, Kun Wei, Cheng Deng, and Muli Yang.
\newblock Siamese contrastive embedding network for compositional zero-shot
  learning.
\newblock In {\em Proceedings of the IEEE/CVF Conference on Computer Vision and
  Pattern Recognition}, pages 9326--9335, 2022.

\bibitem{li2020symmetry}
Yong-Lu Li, Yue Xu, Xiaohan Mao, and Cewu Lu.
\newblock Symmetry and group in attribute-object compositions.
\newblock In {\em Proceedings of the IEEE/CVF Conference on Computer Vision and
  Pattern Recognition}, pages 11316--11325, 2020.

\bibitem{lu2017fully}
Yongxi Lu, Abhishek Kumar, Shuangfei Zhai, Yu Cheng, Tara Javidi, and Rogerio
  Feris.
\newblock Fully-adaptive feature sharing in multi-task networks with
  applications in person attribute classification.
\newblock In {\em Proceedings of the IEEE conference on computer vision and
  pattern recognition}, pages 5334--5343, 2017.

\bibitem{mancini2021open}
Massimiliano Mancini, Muhammad~Ferjad Naeem, Yongqin Xian, and Zeynep Akata.
\newblock Open world compositional zero-shot learning.
\newblock In {\em Proceedings of the IEEE/CVF conference on computer vision and
  pattern recognition}, pages 5222--5230, 2021.

\bibitem{mikolov2013distributed}
Tomas Mikolov, Ilya Sutskever, Kai Chen, Greg~S Corrado, and Jeff Dean.
\newblock Distributed representations of words and phrases and their
  compositionality.
\newblock {\em Advances in neural information processing systems}, 26, 2013.

\bibitem{misra2017red}
Ishan Misra, Abhinav Gupta, and Martial Hebert.
\newblock From red wine to red tomato: Composition with context.
\newblock In {\em Proceedings of the IEEE Conference on Computer Vision and
  Pattern Recognition}, pages 1792--1801, 2017.

\bibitem{naeem2021learning}
Muhammad~Ferjad Naeem, Yongqin Xian, Federico Tombari, and Zeynep Akata.
\newblock Learning graph embeddings for compositional zero-shot learning.
\newblock In {\em Proceedings of the IEEE/CVF Conference on Computer Vision and
  Pattern Recognition}, pages 953--962, 2021.

\bibitem{2018attrasop}
Tushar Nagarajan and Kristen Grauman.
\newblock Attributes as operators: factorizing unseen attribute-object
  compositions.
\newblock In {\em Proceedings of the European Conference on Computer Vision
  (ECCV)}, pages 169--185, 2018.

\bibitem{nair2010rectified}
Vinod Nair and Geoffrey~E Hinton.
\newblock Rectified linear units improve restricted boltzmann machines.
\newblock In {\em Icml}, 2010.

\bibitem{nan2019recognizing}
Zhixiong Nan, Yang Liu, Nanning Zheng, and Song-Chun Zhu.
\newblock Recognizing unseen attribute-object pair with generative model.
\newblock In {\em Proceedings of the AAAI Conference on Artificial
  Intelligence}, volume~33, pages 8811--8818, 2019.

\bibitem{nayak2022learning}
Nihal~V Nayak, Peilin Yu, and Stephen~H Bach.
\newblock Learning to compose soft prompts for compositional zero-shot
  learning.
\newblock {\em arXiv preprint arXiv:2204.03574}, 2022.

\bibitem{purushwalkam2019task}
Senthil Purushwalkam, Maximilian Nickel, Abhinav Gupta, and Marc'Aurelio
  Ranzato.
\newblock Task-driven modular networks for zero-shot compositional learning.
\newblock In {\em Proceedings of the IEEE/CVF International Conference on
  Computer Vision}, pages 3593--3602, 2019.

\bibitem{ruis2021independent}
Frank Ruis, Gertjan Burghouts, and Doina Bucur.
\newblock Independent prototype propagation for zero-shot compositionality.
\newblock {\em Advances in Neural Information Processing Systems},
  34:10641--10653, 2021.

\bibitem{saini2022disentangling}
Nirat Saini, Khoi Pham, and Abhinav Shrivastava.
\newblock Disentangling visual embeddings for attributes and objects.
\newblock In {\em Proceedings of the IEEE/CVF Conference on Computer Vision and
  Pattern Recognition}, pages 13658--13667, 2022.

\bibitem{siddiquie2011image}
Behjat Siddiquie, Rogerio~S Feris, and Larry~S Davis.
\newblock Image ranking and retrieval based on multi-attribute queries.
\newblock In {\em CVPR 2011}, pages 801--808. IEEE, 2011.

\bibitem{singh2016end}
Krishna~Kumar Singh and Yong~Jae Lee.
\newblock End-to-end localization and ranking for relative attributes.
\newblock In {\em European Conference on Computer Vision}, pages 753--769.
  Springer, 2016.

\bibitem{srivastava2014dropout}
Nitish Srivastava, Geoffrey Hinton, Alex Krizhevsky, Ilya Sutskever, and Ruslan
  Salakhutdinov.
\newblock Dropout: a simple way to prevent neural networks from overfitting.
\newblock {\em The journal of machine learning research}, 15(1):1929--1958,
  2014.

\bibitem{su2016deep}
Chi Su, Shiliang Zhang, Junliang Xing, Wen Gao, and Qi Tian.
\newblock Deep attributes driven multi-camera person re-identification.
\newblock In {\em European conference on computer vision}, pages 475--491.
  Springer, 2016.

\bibitem{wang2019tafe}
Xin Wang, Fisher Yu, Ruth Wang, Trevor Darrell, and Joseph~E Gonzalez.
\newblock Tafe-net: Task-aware feature embeddings for low shot learning.
\newblock In {\em Proceedings of the IEEE/CVF conference on computer vision and
  pattern recognition}, pages 1831--1840, 2019.

\bibitem{xu2021zero}
Guangyue Xu, Parisa Kordjamshidi, and Joyce~Y Chai.
\newblock Zero-shot compositional concept learning.
\newblock {\em arXiv preprint arXiv:2107.05176}, 2021.

\bibitem{xu2021relation}
Ziwei Xu, Guangzhi Wang, Yongkang Wong, and Mohan~S Kankanhalli.
\newblock Relation-aware compositional zero-shot learning for attribute-object
  pair recognition.
\newblock {\em IEEE Transactions on Multimedia}, 2021.

\bibitem{yang2020learning}
Muli Yang, Cheng Deng, Junchi Yan, Xianglong Liu, and Dacheng Tao.
\newblock Learning unseen concepts via hierarchical decomposition and
  composition.
\newblock In {\em Proceedings of the IEEE/CVF Conference on Computer Vision and
  Pattern Recognition}, pages 10248--10256, 2020.

\bibitem{yang2022decomposable}
Muli Yang, Chenghao Xu, Aming Wu, and Cheng Deng.
\newblock A decomposable causal view of compositional zero-shot learning.
\newblock {\em IEEE Transactions on Multimedia}, 2022.

\bibitem{yu2014fine}
Aron Yu and Kristen Grauman.
\newblock Fine-grained visual comparisons with local learning.
\newblock In {\em Proceedings of the IEEE conference on computer vision and
  pattern recognition}, pages 192--199, 2014.

\bibitem{yu2017semantic}
Aron Yu and Kristen Grauman.
\newblock Semantic jitter: Dense supervision for visual comparisons via
  synthetic images.
\newblock In {\em Proceedings of the IEEE International Conference on Computer
  Vision}, pages 5570--5579, 2017.

\bibitem{zhang2022learning}
Tian Zhang, Kongming Liang, Ruoyi Du, Xian Sun, Zhanyu Ma, and Jun Guo.
\newblock Learning invariant visual representations for compositional zero-shot
  learning.
\newblock {\em arXiv preprint arXiv:2206.00415}, 2022.

\bibitem{zhang2019adacos}
Xiao Zhang, Rui Zhao, Yu Qiao, Xiaogang Wang, and Hongsheng Li.
\newblock Adacos: Adaptively scaling cosine logits for effectively learning
  deep face representations.
\newblock In {\em Proceedings of the IEEE/CVF Conference on Computer Vision and
  Pattern Recognition}, pages 10823--10832, 2019.

\end{thebibliography}
}

\end{document}